\documentclass{article} 
\usepackage[dvipsnames]{xcolor}
\usepackage{iclr2022_conference,times}
\usepackage{mathtools} \usepackage{multirow}
\usepackage{booktabs}
\usepackage{colortbl}
\usepackage{subfigure}
\usepackage[export]{adjustbox}
\usepackage{afterpage}
\usepackage{wrapfig}
\usepackage[ruled,linesnumbered]{algorithm2e} 
\usepackage{newtxtext}
\usepackage{fancyvrb}
\usepackage[frozencache,cachedir=.]{minted}
\usepackage{tabularx}
\usepackage{makecell}

\usepackage{amsmath,amsfonts,bm}

\def\eqref#1{equation~\ref{#1}}

\def\1{\bm{1}}

\DeclareMathAlphabet{\mathsfit}{\encodingdefault}{\sfdefault}{m}{sl}
\SetMathAlphabet{\mathsfit}{bold}{\encodingdefault}{\sfdefault}{bx}{n}

\usepackage{xspace}
\newcolumntype{Y}{>{\centering\arraybackslash}X}
\newcommand{\ie}{{\emph{i.e.}},\xspace}

\newcommand{\eg}{{\emph{e.g.}},\xspace}

\newcommand{\cin}{c_{\mathsf{in}}}

\newcommand{\coldot}[1][black]{\textcolor{#1}{\ensuremath\bullet}}

\usepackage{amsthm}

\theoremstyle{remark}

\theoremstyle{definition}

\usepackage{soul}
\definecolor{almond}{rgb}{0.94, 0.87, 0.8}
\definecolor{antiquewhite}{rgb}{0.98, 0.92, 0.84}
\definecolor{electriclavender}{rgb}{0.96, 0.73, 1.0}
\definecolor{cloudgray}{rgb}{0.92, 0.92, 0.92}
\definecolor{dawnblue}{rgb}{0.84, 0.92, 1.0}
\definecolor{rdd}{rgb}{0.85, 0.8, 0.95}
\newcommand{\hlc}[2][pink]{{\sethlcolor{#1}\hl{#2}}}
\newcommand{\result}[1]{\hlc[dawnblue]{#1}}

\definecolor{cb2}{RGB}{146,   0,   0}
\definecolor{cb3}{RGB}{182, 109, 255}
\definecolor{cb1}{RGB}{ 0, 109, 219}
\definecolor{cb4}{RGB}{ 36, 255,  36}

\usepackage{color, colortbl}
\definecolor{Gray}{gray}{0.9}
\usepackage{hyperref}
 \hypersetup{
     colorlinks=true,
     linkcolor=MidnightBlue,
     filecolor=MidnightBlue,
     citecolor = MidnightBlue,
     urlcolor=MidnightBlue,
 }

\usepackage{url}

\usepackage{ifthen}
\usepackage{amssymb}
\newboolean{showcomments}
\setboolean{showcomments}{true} 
\ifthenelse{\boolean{showcomments}}
  {\newcommand{\nb}[2]{
    \fcolorbox{gray}{yellow}{\bfseries\sffamily\scriptsize#1}
    {\sf\small$\blacktriangleright$\textit{#2}$\blacktriangleleft$}
   }
   
  }
  {\newcommand{\nb}[2]{}
   
  }

\newcommand{\todonr}[1]{\smash{\fcolorbox{gray}{red}{?}}}

\title{\includegraphics[height=0.72em]{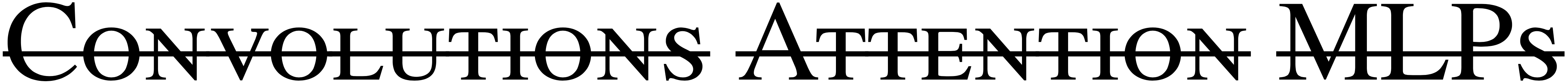} \\ Patches Are All You Need? \includegraphics[height=1.2em, trim={10 10 10 0},clip]{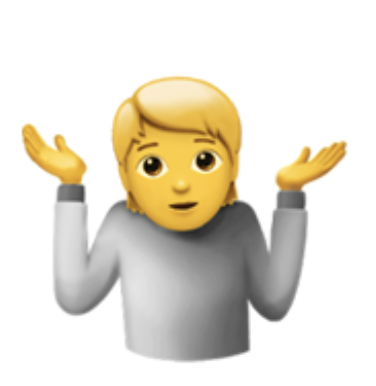}}

\author{%
	Asher Trockman,\; J. Zico Kolter$^1$ \\
	Carnegie Mellon University  and $^1$Bosch Center for AI \\
}

\iclrfinalcopy
\begin{document}

\maketitle

\begin{abstract}
	Although convolutional networks have been the dominant architecture for vision tasks for many years, recent experiments have shown that Transformer-based models, most notably the Vision Transformer (ViT), may exceed their performance in some settings. However, due to the quadratic runtime of the self-attention layers in Transformers, ViTs require the use of patch embeddings, which group together small regions of the image into single input features, in order to be applied to larger image sizes. This raises a question: Is the performance of ViTs due to the inherently-more-powerful Transformer architecture, or is it at least partly due to using patches as the input representation? In this paper, we present some evidence for the latter: specifically, we propose the ConvMixer, an extremely simple model that is similar in spirit to the ViT and the even-more-basic MLP-Mixer in that it operates directly on patches as input, separates the mixing of spatial and channel dimensions, and maintains equal size and resolution throughout the network. In contrast, however, the ConvMixer uses only standard convolutions to achieve the mixing steps. Despite its simplicity, we show that the ConvMixer outperforms the ViT, MLP-Mixer, and some of their variants for similar parameter counts and data set sizes, in addition to outperforming classical vision models such as the ResNet. Our code is available at \url{https://github.com/locuslab/convmixer}.
\end{abstract}

\section{Introduction}

\begin{wrapfigure}[15]{r}{0.4\textwidth}
 \vspace*{-1.25em}
  \begin{center}
	  \includegraphics[width=0.4\textwidth,trim={0 0 0 0},clip]{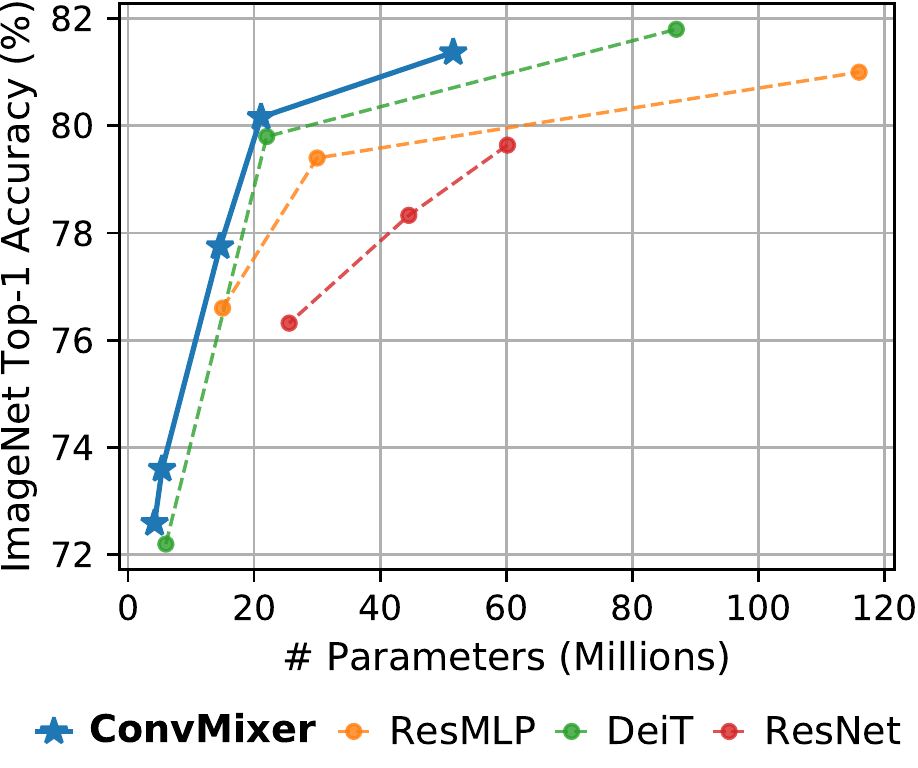}
  \end{center}
	\vspace*{-1.75em}
	\caption{Accuracy \emph{vs.} parameters, trained and evaluated on ImageNet-1k.}
  \label{patch-embeddings}
\end{wrapfigure}

\looseness=-1
For many years, convolutional neural networks have been the dominant architecture for deep learning systems
applied to computer vision tasks.
But recently, architectures based upon \emph{Transformer} models,
\eg the so-called Vision Transformer architecture~\citep{vit},
have demonstrated compelling performance in many of these tasks,
often outperforming classical convolutional architectures, especially for large data sets. 
An understandable assumption, then, is that it is only a matter of time before Transformers become 
the dominant architecture for vision domains,
just as they have for language processing.
In order to apply Transformers to images, however, the representation had to be changed:
because the computational cost of the self-attention layers used in Transformers would
scale quadratically with the number of pixels per image if applied naively at the per-pixel level, the compromise
was to first split the image into multiple 
``patches'', linearly embed them, and then apply the transformer directly to this collection of patches.

In this work, we explore the question of whether, fundamentally,
the strong performance of vision transformers may result more from this
patch-based representation than from the Transformer architecture itself.
We develop a very simple convolutional architecture
which we dub the ``ConvMixer'' due to its similarity to the recently-proposed MLP-Mixer~\citep{mixer}.
This architecture is similar to the Vision Transformer (and MLP-Mixer) in many respects:
it directly operates on patches, it maintains an equal-resolution-and-size
representation throughout all layers, it does no downsampling of the representation at successive layers,
and it separates ``channel-wise mixing'' from the ``spatial mixing'' of information.
But unlike the Vision Transformer and MLP-Mixer, our architecture does
all these operations via only standard convolutions.

\looseness=-1
The chief result we show in this paper is that this ConvMixer architecture,
despite its extreme simplicity (it can be implemented in $\approx6$ lines of dense PyTorch code),
outperforms both ``standard'' computer vision models such as ResNets of similar parameter counts
\emph{and} some corresponding Vision Transformer and MLP-Mixer variants,
even with a slate of additions intended to make those architectures more performant on smaller data sets.
Importantly, this is despite the fact that \emph{we did not design our experiments to maximize accuracy nor speed,
in contrast to the models we compared against.
Our results suggest that,} at least to some extent, the \emph{patch representation itself}
may be a critical component to the ``superior'' performance of newer architectures like Vision Transformers.
While these results are naturally just a snapshot, and more experiments are required to exactly disentangle
the effect of patch embeddings from other factors, we believe that this provides a strong
``convolutional-but-patch-based'' baseline to compare against for more advanced architectures in the future.  

\section{A Simple Model: ConvMixer}

\begin{figure}[t!]
	\centering
	\includegraphics[width=0.95\textwidth,trim={5 30 18 30},clip]{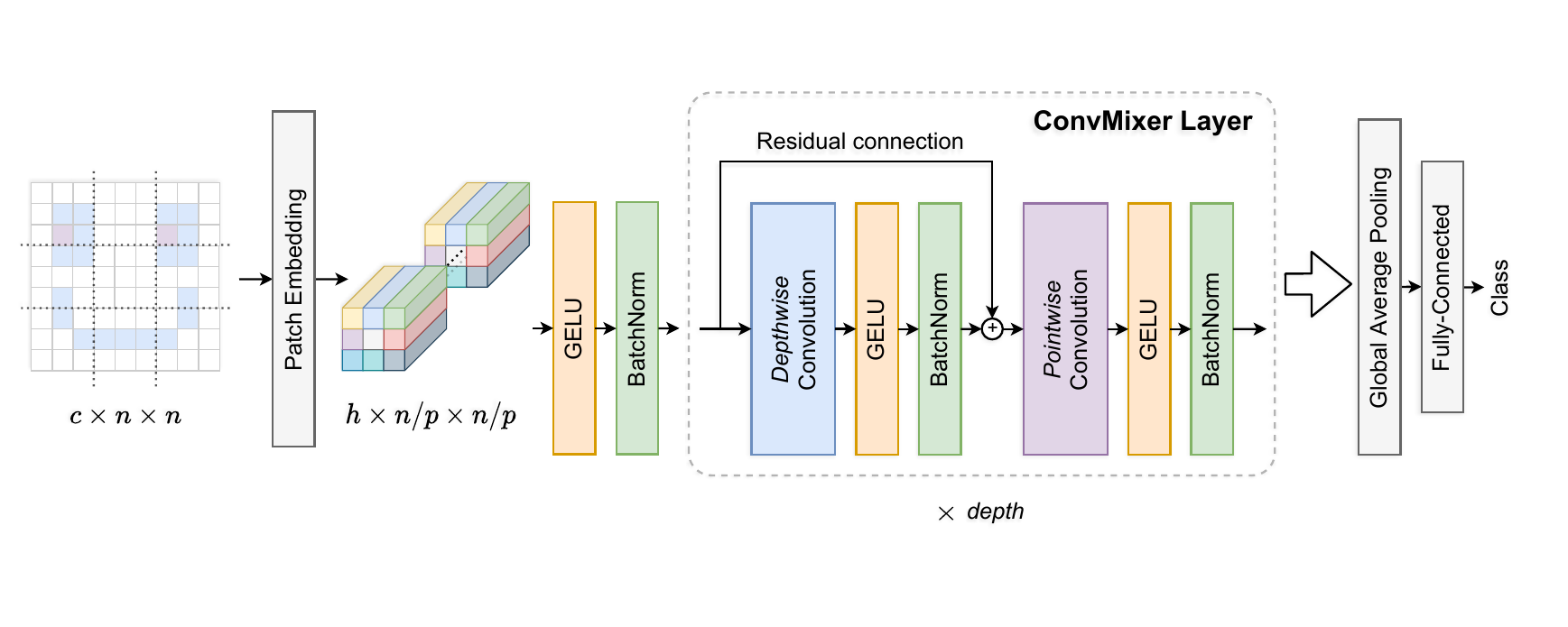}
	\vspace*{-1em}
	\caption{ConvMixer uses ``tensor layout'' patch embeddings to preserve locality, and then applies $d$ copies of a simple fully-convolutional block consisting of \emph{large-kernel} depthwise convolution followed by pointwise convolution, before finishing with global pooling and a simple linear classifier.}
	\label{fig-cmix}
\end{figure}

Our model, dubbed \emph{ConvMixer}, consists of a patch embedding layer followed by repeated
applications of a simple fully-convolutional block. We maintain the spatial structure
of the patch embeddings, as illustrated in Fig.~\ref{fig-cmix}.
Patch embeddings with patch size $p$ and embedding dimension $h$
can be implemented as convolution with $\cin$ input channels, $h$ output channels, kernel size $p$, and stride $p$:%
\begin{equation}
	z_0 = \mathsf{BN}\left(\sigma\{\text{\texttt{Conv}}_{\cin \to h}(X, \text{\texttt{stride=}}p, \text{\texttt{kernel\_size}=}p)\}\right)
\end{equation}
The ConvMixer block itself consists of depthwise convolution (\ie grouped convolution with groups equal to the number of channels, $h$)
followed by pointwise (\ie kernel size $1\times1$) convolution. 
As we will explain in Sec.~\ref{sec-experiments}, ConvMixers work best
with unusually large kernel sizes for the depthwise convolution.
Each of the convolutions
is followed by an activation and post-activation BatchNorm:%
\begin{align}
	z_l' &= \mathsf{BN}\left(\sigma\{\text{\texttt{ConvDepthwise}}(z_{l-1})\}\right) + z_{l-1} \label{eq-depthwise} \\
	z_{l+1} &= \mathsf{BN}\left(\sigma\{\text{\texttt{ConvPointwise}}(z_l')\}\right) \label{eq-pointwise}
\end{align}
\looseness=-1
After many applications of this block, we perform global pooling to get a feature vector of size $h$,
which we pass to a softmax classifier.
See Fig.~\ref{pytorch-code} for an implementation of ConvMixer in PyTorch.

\begin{figure}[t!]
\centering
{\footnotesize
\begin{minted}[linenos]{python}
def ConvMixer(h, depth, kernel_size=9, patch_size=7, n_classes=1000):
    Seq, ActBn = nn.Sequential, lambda x: Seq(x, nn.GELU(), nn.BatchNorm2d(h))
    Residual = type('Residual', (Seq,), {'forward': lambda self, x: self[0](x) + x})
    return Seq(ActBn(nn.Conv2d(3, h, patch_size, stride=patch_size)),
        *[Seq(Residual(ActBn(nn.Conv2d(h, h, kernel_size, groups=h, padding="same"))),
          ActBn(nn.Conv2d(h, h, 1))) for i in range(depth)],
         nn.AdaptiveAvgPool2d((1,1)), nn.Flatten(), nn.Linear(h, n_classes))
\end{minted}
}
	\vspace*{-1em}
	\caption{Implementation of ConvMixer in PyTorch; see Appendix~\ref{apx-implementation} for more implementations.}
	\label{pytorch-code}
	\vspace*{-1em}
\end{figure}

\looseness=-1
\textbf{Design parameters.} An instantiation of ConvMixer depends on four parameters:
(1) the ``width'' or hidden dimension $h$ (\ie the dimension of the patch embeddings),
(2) the depth $d$, or the number of repetitions of the ConvMixer layer,
(3) the patch size $p$ which controls the internal resolution of the model,
(4) the kernel size $k$ of the depthwise convolutional layer.
We name ConvMixers after their hidden dimension and depth, like ConvMixer-$h/d$.
We refer to the original input size $n$ divided by the patch size $p$ as the
\emph{internal resolution}; note, however, that ConvMixers support variable-sized inputs.

\textbf{Motivation.} Our architecture is based on the idea of \emph{mixing},
as in~\cite{mixer}. In particular, we chose depthwise convolution to mix \emph{spatial locations}
and pointwise convolution to mix \emph{channel locations}.
A key idea from previous work is that MLPs and self-attention can mix
distant spatial locations, \ie they can have an arbitrarily large receptive field.
Consequently, we used convolutions with an unusually large kernel size to mix distant spatial locations.

While self-attention and MLPs are theoretically more flexible, allowing
for large receptive fields and content-aware behavior,
the inductive bias of convolution is well-suited to vision tasks and leads to high data efficiency.
By using such a standard operation, we also get a glimpse into the effect of the patch representation itself
in contrast to the conventional pyramid-shaped, progressively-downsampling design of convolutional networks.

\begin{table}[t!]
\begin{center}

\begin{tabularx}{\textwidth}{l|c|c|c|c|c|c|Y}
\toprule
\multicolumn{8}{c}{Current ``Most Interesting'' \textbf{ConvMixer} Configurations \emph{vs.} Other Simple Models} \\
\midrule
\thead{Network} &
\thead{Patch \\ Size} &
\thead{Kernel \\ Size} &
\thead{\# Params\\($\times10^6$)} &
\thead{Throughput\\(img/sec)} & 
\thead{Act. \\ Fn.} &
\thead{\# Epochs} &
\thead{ImNet \\top-1 (\%)} \\
\midrule
ConvMixer-1536/20                     & 7   &9& 51.6  & 134  &G& 150 & 81.37 \\
ConvMixer-768/32                       & 7   &7& 21.1  & 206 &R& 300 & 80.16 \\ 
\midrule
ResNet-152                            & --  &3& 60.2 & 828 &R& 150 & 79.64 \\
DeiT-B                                & 16  &--&  86 & 792  &G&  300 & 81.8 \\
ResMLP-B24/8                          & 8   &--&129  & 181  &G& 400 &  81.0 \\
\bottomrule
\end{tabularx}
	\caption{Models trained and evaluated on $224\times224$ ImageNet-1k only. See more in Appendix~\ref{other-models}.}
\vspace*{-1.5em}
\label{convmixers}
\end{center}
\end{table}

\section{Experiments}
\label{sec-experiments}

\textbf{Training setup.} We primarily evaluate ConvMixers on ImageNet-1k classification without any pretraining or additional data.
We added ConvMixer to the \texttt{timm} framework~\citep{timm} and trained it with nearly-standard settings:
we used RandAugment~\citep{randaug}, mixup~\citep{mixup}, CutMix~\citep{cutmix}, random erasing~\citep{random-erasing},
and gradient norm clipping in addition to default \texttt{timm} augmentation.
We used the AdamW~\citep{adamw} optimizer and a simple triangular learning rate schedule.
Due to limited compute, \emph{we did absolutely no hyperparameter tuning} on ImageNet and trained for fewer epochs than competitors.
Consequently, our models could be over- or under-regularized, and
the accuracies we report likely underestimate the capabilities of our model.

\textbf{Results.} A ConvMixer-1536/20 with 52M parameters can achieve 81.4\% top-1 accuracy
on ImageNet, and a ConvMixer-768/32 with 21M parameters 80.2\% (see Table~\ref{convmixers}). 
Wider ConvMixers seem to converge in fewer epochs, but are memory- and compute-hungry.
They also work best with large kernel sizes:
ConvMixer-1536/20 lost $\approx1\%$ accuracy when reducing the kernel size from $k=9$ to $k=3$
(we discuss kernel sizes more in Appendix~\ref{other-models} \& \ref{apx-cifar}).
ConvMixers with smaller patches are substantially better in our experiments, similarly to~\cite{isomobile};
we believe larger patches require deeper ConvMixers.
With everything held equal except increasing the patch size from 7 to 14,
ConvMixer-1536/20 achieves 78.9\% top-1 accuracy but is around 4$\times$ faster.
We trained one model with ReLU to demonstrate that GELU~\citep{gelu},
which is popular in recent isotropic models, isn't necessary.

\looseness=-1
\textbf{Comparisons.}
Our model and ImageNet1k-only training setup closely resemble that of
recent patch-based models like DeiT~\citep{deit}.
Due to ConvMixer's simplicity,
we focus on comparing to only the most basic isotropic patch-based architectures
adapted to the ImageNet-1k setting, namely DeiT and ResMLP.
Attempting a fair comparison with a standard baseline,
we trained ResNets using exactly the same parameters as ConvMixers;
while this choice of parameters is suboptimal~\citep{strikesback}, it is 
likely also suboptimal for ConvMixers, since we did \emph{no hyperparameter tuning}.

\looseness=-1
Looking at Table~\ref{convmixers} and Fig.~\ref{patch-embeddings},
ConvMixers achieve competitive accuracies for a given parameter budget:
ConvMixer-1536/20 outperforms both ResNet-152 and ResMLP-B24 despite
having substantially fewer parameters and is competitive with DeiT-B.
ConvMixer-768/32 uses just a third of the parameters of ResNet-152, but is similarly accurate.
Note that unlike ConvMixer, the DeiT and ResMLP results involved hyperparameter tuning,
and when substantial resources are dedicated to tuning ResNets, including training for twice as many epochs,
they only outperform an equivalently-sized ConvMixer by $\approx0.2\%$~\citep{strikesback}.
However, ConvMixers are substantially slower at inference than the competitors,
likely due to their smaller patch size;
hyperparameter tuning and optimizations could narrow this gap.
For more discussion and comparisons, see Table~\ref{models} and Appendix~\ref{other-models}.

\textbf{Hyperparameters.} For experiments presented in the main text,
we used only one set of ``common sense'' parameters for the regularization methods.
Recently, we adapted parameters from the A1 procedure in \citet{strikesback}, published after our work,
which were better than our initial guess, \eg giving $+0.8\%$ for ConvMixer-1536/20, or \result{82.2\% top-1 accuracy}
(see Appendix~\ref{other-models}).

\textbf{CIFAR-10 Experiments.} We also performed smaller-scale experiments on CIFAR-10, where
ConvMixers achieve over 96\% accuracy with as few as 0.7M parameters, demonstrating the data 
efficiency of the convolutional inductive bias.  Details of these experiments are presented in 
Appendix~\ref{apx-cifar}.

\section{Related Work}
\label{related-work}
\looseness=-1
\textbf{Isotropic architectures.}
Vision transformers have inspired a new paradigm of ``isotropic'' architectures,
\ie those with equal size and shape throughout the network,
which use patch embeddings for the first layer.
These models look similar to repeated transformer-encoder blocks~\citep{vaswani2017attention}
with different operations replacing the self-attention and MLP operations.
For example, MLP-Mixer~\citep{mixer} replaces them both with MLPs applied across
different dimensions (\ie spatial and channel location mixing);
ResMLP~\citep{resmlp} is a data-efficient variation on this theme.
CycleMLP~\citep{chen2021cyclemlp}, gMLP~\citep{liu2021pay}, and vision permutator~\citep{permutator},
replace one or both blocks with various novel operations.
These are all quite performant, which is typically attributed to the novel choice of operations.
In contrast, \cite{dyena} proposed an MLP-based isotropic vision model, and also hypothesized patch embeddings
could be behind its performance.
ResMLP tried replacing its linear interaction layer with (small-kernel) convolution and achieved good performance, 
but kept its MLP-based cross-channel layer and did not explore convolutions further.
As our investigation of ConvMixers suggests, these works may conflate
the effect of the new operations (like self-attention and MLPs) with the effect of the use of patch embeddings
and the resulting isotropic architecture.

\looseness=-1
A study predating vision transformers investigate
isotropic (or ``isometric'') MobileNets~\citep{isomobile}, and even implements
patch embeddings under another name.
Their architecture simply repeats an isotropic MobileNetv3 block.
They identify a tradeoff between patch size and accuracy that matches our experience,
and train similarly performant models (see Appendix~\ref{other-models}, Table~\ref{models}).
However, their block is substantially more complex than ours; simplicity and motivation sets our work apart.

\looseness=-1
\textbf{Patches aren't all you need.}
Several papers have increased vision transformer performance by replacing standard patch embeddings
with a different stem: \citet{xiao2021early} and \citet{yuan2021incorporating} use a standard convolutional stem,
while \citet{yuan2021tokens} repeatedly combines nearby patch embeddings.
However, this conflates the effect of using patch embeddings with the effect of adding
convolution or similar inductive biases \eg locality. We attempt to focus on the use of patches.

\textbf{CNNs meet ViTs.}
Many efforts have been made to incorporate features of convolutional networks
into vision transformers and vice versa.
Self-attention can emulate convolution~\citep{cordonnier2019relationship}
and can be initialized or regularized to be like it~\citep{d2021convit};
other works simply add convolution operations to transformers~\citep{dai2021coatnet, guo2021cmt},
or include downsampling to be more like
traditional pyramid-shaped convolutional networks~\citep{wang2021pyramid}.
Conversely, self-attention or attention-like operations can supplement
or replace convolution in ResNet-style models~\citep{bello2019attention, ramachandran2019stand, bello2021lambdanetworks}.
While all of these attempts have been successful in one way or another, 
they are orthogonal to this work, which aims to emphasize the effect
of the architecture common to most ViTs by showcasing it with a less-expressive operation.

\section{Conclusion}

\looseness=-1
We presented ConvMixers, an extremely simple class of models that independently
mixes the spatial and channel locations of patch embeddings using only standard convolutions.
We also highlighted that using large kernel sizes, inspired by the large receptive fields of ViTs and MLP-Mixers,
provides a substantial performance boost.
While neither our model nor our experiments were designed to maximize accuracy or speed,
\ie we did not search for good hyperparameters,
ConvMixers outperform the Vision Transformer and MLP-Mixer, and
are competitive with ResNets, DeiTs, and ResMLPs.

We provided evidence that the
increasingly common ``isotropic'' architecture with a simple patch embedding stem
is itself a powerful template for deep learning.
Patch embeddings allow all the downsampling to happen at once,
immediately decreasing the internal resolution and
thus increasing the effective receptive field size,
making it easier to mix distant spatial information.
Our title, while an exaggeration, points out that attention isn't
the only export from language processing into computer vision: tokenizing inputs, \ie
using patch embeddings, is also a powerful and important takeaway.

\looseness=-1
While our model is not state-of-the-art, we find its simple patch-mixing design to be compelling.
We hope that ConvMixers can serve as a baseline for future patch-based architectures
with novel operations, or that they can provide a basic template for new conceptually simple and performant models.

\textbf{Future work.}
We are optimistic that a deeper ConvMixer with larger patches
could reach a desirable tradeoff between accuracy, parameters, and throughput
after longer training and more regularization and hyperparameter tuning,
similarly to how \cite{strikesback} enhanced ResNet performance through
carefully-designed training regimens.
Low-level optimization of large-kernel depthwise convolution could substantially
increase throughput, and small enhancements to our architecture like the addition of bottlenecks
or a more expressive classifier could trade simplicity for performance.

\looseness=-1
Due to its large internal resolution and isotropic design, ConvMixer
may be especially well-suited for semantic segmentation, and it would be useful to run experiments on this task
with a ConvMixer-like model and on other tasks such as object detection.
More experiments could be designed to more clearly extricate the effect of patch embeddings
from other architectural choices.
In particular, for a more in-depth comparison to ViTs and MLP-Mixers, which excel when trained on very large data sets,
it is important to investigate the performance of ConvMixers in the regime of large-scale pre-training.

\bibliography{references}
\bibliographystyle{iclr2022_conference}

\newpage
\appendix

\section{Comparison to other models}
\label{other-models}

\begin{table}[ht]
\begin{center}
\begin{tabularx}{\textwidth}{l|c|c|c|c|c|c|c}
\toprule
\multicolumn{8}{c}{Comparison with other simple models trained on \textbf{ImageNet-1k only} with input size 224.} \\
\midrule
\thead{Network} &
\thead{Patch \\ Size} &
\thead{Kernel \\ Size} &
\thead{\# Params\\($\times10^6$)} &
\thead{Throughput\\(img/sec)} & 
\thead{Act. \\ Fn.} &
\thead{\# Epochs} &
\thead{ImNet \\top-1 (\%)} \\
\midrule
ConvMixer-1536/20$^\bigstar$                  & 7   &9& 51.6  & 134  &G& 150 & 82.20 \\
ConvMixer-1536/20 \coldot[cb1]                & 7   &9& 51.6  & 134  &G& 150 & 81.37 \\
ConvMixer-1536/20$^\bigstar$                  & 7   &3& 49.4  & 246 &G& 150 & 81.60 \\
ConvMixer-1536/20                             & 7   &3& 49.4  & 246 &G& 150 & 80.43 \\
ConvMixer-1536/20                             & 14  &9& 52.3  & 538 &G& 150 & 78.92 \\
ConvMixer-1536/24$^\bigstar$                  & 14  &9& 62.3  & 447 &G& 150 & 80.21 \\
ConvMixer-768/32 \coldot[cb2]                 & 7   &7& 21.1  & 206 &R& 300 & 80.16 \\
ConvMixer-1024/16                             & 7   &9& 19.4  & 244 &G& 100 & 79.45 \\
ConvMixer-1024/12                             & 7   &8& 14.6  & 358 &G& 90 & 77.75 \\ 
ConvMixer-512/16                              & 7   &8& 5.4   & 599 &G& 90 & 73.76 \\
ConvMixer-512/12 \coldot[cb4]                 & 7   &8& 4.2   & 798 &G& 90 & 72.59 \\
ConvMixer-768/32                              & 14  &3& 20.2  & 1235&R& 300 & 74.93 \\
ConvMixer-1024/20 \coldot[cb3]                & 14  &9& 24.4  & 750 &G& 150 & 76.94 \\
\midrule
ResNet-152 \coldot[cb1]                       & --  &3& 60.2 & 828 &R& 150 & 79.64 \\
ResNet-101 \coldot[cb3]                       & --  &3& 44.6 & 1187&R& 150 & 78.33 \\
ResNet-50                                     & --  &3& 25.6 & 1739&R& 150 & 76.32 \\
\midrule
DeiT-B$^\dagger$                              & 7   &--&  86.7 & 83  &G& --  &  -- \\
DeiT-S$^\dagger$                              & 7   &--&  22.1 & 174 &G&--  &  -- \\
DeiT-Ti$^\dagger$                             & 7   &--& 5.7   & 336 &G& --  &  -- \\
DeiT-B \coldot[cb1]                           & 16  &--&  86 & 792  &G&  300 & 81.8 \\
DeiT-S \coldot[cb3]                           & 16  &--&  22 & 1610 &G& 300 &  79.8 \\
DeiT-Ti \coldot[cb4]                          & 16  &--& 5.7   & 2603 &G& 300 & 72.2 \\
\midrule
ResMLP-S12/8 \coldot[cb2]                     & 8   &--&22.1 & 872  &G& 400 &  79.1 \\
ResMLP-B24/8 \coldot[cb1]                     & 8   &--&129   & 181  &G& 400 &  81.0 \\
ResMLP-B24                                    & 16  &--& 116 & 1597 &G& 400 &  81.0 \\
\midrule
Swin-S \coldot[cb1]                           & 4 & -- & 50 & 576 & G & 300 & 83.0 \\
Swin-T \coldot[cb2] 			              & 4 & -- & 29 & 878 & G & 300 & 81.3 \\
\midrule
ViT-B/16 \coldot[cb1]                         & 16  &--&  86 & 789  &G& 300 & 77.9 \\
\midrule
Mixer-B/16 \coldot[cb1]                       & 16  &--&  59 & 1025  &G& 300 & 76.44 \\
\midrule
Isotropic MobileNetv3 \coldot[cb2]            & 8   &3& 20  & --   &R& --& 80.6\\
Isotropic MobileNetv3  \coldot[cb3]           & 16  &3& 20  & --   &R&  --& 77.6\\
\bottomrule
\end{tabularx}
\looseness=-1
\caption{Throughputs measured on an RTX8000 GPU using batch size 64 and fp16. ConvMixers and ResNets trained ourselves.
	Other statistics: DeiT~\citep{deit}, ResMLP~\citep{resmlp}, Swin~\citep{swin}, ViT~\citep{vit}, MLP-Mixer~\citep{mixer}, 
    Isotropic MobileNets~\citep{isomobile}. We think models with matching colored dots (\coldot[black]) are informative to compare with each other. $^\dagger$Throughput tested, but not trained. Activations: \textbf{R}eLU, \textbf{G}ELU.\newline $^\bigstar$Using new, better regularization hyperparameters based on \citet{strikesback}'s A1 procedure.}
\label{models}
\end{center}
\end{table}

\textbf{Experiment overview.}
We did not design our experiments to maximize accuracy:
We chose ``common sense'' parameters for \texttt{timm} and its augmentation settings,
found that it worked well for a ConvMixer-1024/12, and stuck with them for the proceeding experiments.
We admit this is not an optimal strategy, however, we were aware from our early experiments on CIFAR-10
that results seemed robust to various small changes.
We did not have access to sufficient compute to attempt to tune hyperparameters for each model:
\eg larger ConvMixers could probably benefit from more regularization than we chose,
and smaller ones from less regularization.
Keeping the parameters the same across ConvMixer instances seemed more reasonable than guessing for each.

\looseness=-1
However, to some extent, we changed the number of epochs per model: for earlier experiments, we merely
wanted a ``proof of concept'', and used only 90--100 epochs. 
Once we saw potential, we increased this to 150 epochs and trained some larger models,
namely ConvMixer-1024/20 with $p=14$ patches and ConvMixer-1536/20
with $p=7$ patches. Then, believing that we should explore deeper-but-less-wide
ConvMixers, and knowing from CIFAR-10 that the deeper models converged more slowly,
we trained ConvMixer-768/32s with $p=14$ and $p=7$ for 300 epochs.
Of course, training time was a consideration: ConvMixer-1536/20 took about 9 days to train (on $10\times$ RTX8000s)
150 epochs, and ConvMixer-768/32 is over twice as fast, making 300 epochs more feasible.

If anything, we believe that in the worst case, the lack of parameter tuning in our experiments
resulted in underestimating the accuracies of ConvMixers.
Further, due to our limited compute and the fact that large models (particularly ConvMixers)
are expensive to train on large data sets, we generally trained our models for fewer
epochs than competition like DeiT and ResMLP (see Table~\ref{models}).

In this revision, we have added some additional results (denoted with a $\bigstar$ in Table~\ref{models})
using hyperparameters loosely based on the precisely-crafted ``A1 training procedure'' from \citet{strikesback}.
In particular, we adjusted parameters for RandAug, Mixup, CutMix, Random Erasing, and weight decay to match
those in the procedure. Importantly, we still only trained for 150 epochs, rather than the 600 epochs used in \citet{strikesback},
and we did not use binary cross-entropy loss nor repeated augmentation.
While we do not think optimal hyperparameters for ResNet would also be optimal for ConvMixer,
these settings are significantly better than the ones we initially chose.
This further highlights the capabilities of ConvMixers, and we are optimistic that further tuning
could lead to still-better performance.
Throughout the paper, we still refer to ConvMixers trained using our initial ``one shot'' selection of hyperparameters.

\textbf{A note on throughput.}
We measured throughput using batches of 64 images in half precision
on a single RTX8000 GPU, averaged over 20 such batches.
In particular, we measured CUDA execution time rather than ``wall-clock'' time.
We noticed discrepancies in the relative throughputs of models, \eg
\cite{deit} reports that ResNet-152 is $2\times$ faster than
DeiT-B, but our measurements show that the two models have nearly the same throughput.
We therefore speculate that our throughputs may underestimate the performance of ResNets
and ConvMixers relative to the transformers.
The difference may be due to using RTX8000 rather than V100 GPUs,
or other low-level differences. Our throughputs were similar for batch sizes 32 and 128.

\textbf{ResNets.} As a simple baseline to which to compare ConvMixers,
we trained three standard ResNets using exactly the same training setup and parameters as ConvMixer-1536/20.
Despite having fewer parameters and being architecturally much simpler,
ConvMixers substantially outperform these ResNets in terms of accuracy.
A possible confounding factor is that ConvMixers use GELU, which may boost performance, while ResNets use ReLU.
In an attempt to rule out this confound, we used ReLU in a later ConvMixer-768/32 experiment
and found that it still achieved competitive accuracy.
We also note that the choice of ReLU \emph{vs.} GELU was not important on CIFAR-10 experiments (see Table~\ref{cifar-ablate}).
However, ConvMixers do have substantially less throughput.

\textbf{DeiTs.} We believe that DeiT is the most reasonable comparison in terms of vision transformers:
It only adds additional regularization, as opposed to architectural additions in the case of CaiT~\citep{cait},
and is then essentially a ``vanilla'' ViT modulo the distillation token (we don't consider distilled architectures).
In terms of a fixed parameter budget, ConvMixers generally outperform DeiTs.
For example, ConvMixer-1536/20 is only 0.43\% less accurate than DeiT-B despite having over 30M fewer parameters;
ConvMixer-768/32 is 0.36\% more accurate than DeiT-S despite having 0.9M fewer parameters;
and ConvMixer-512/16 is 0.39\% more accurate than DeiT-Ti for nearly the same number of parameters.
Admittedly, none of the ConvMixers are very competitive in terms of throughput, with the closest being
the ConvMixer-512/16 which is $4\times$ slower than DeiT-Ti.

\looseness=-1
A confounding factor is the difference in patch size between DeiT and ConvMixer;
DeiT uses $p=16$ while ConvMixer uses $p=7$. This means DeiT is substantially faster.
However, ConvMixers using larger patches are not as competitive.
While we were not able to train DeiTs with larger patch sizes, it is possible that they would
outperform ConvMixers on the parameter count \emph{vs.} accuracy curve;
however, we tested their throughput for $p=7$, and they are even slower than ConvMixers.
Given the difference between convolution and self-attention, we are not sure it is
salient to control for patch size differences.

\looseness=-1
DeiTs were subject to more hyperparameter tuning than ConvMixers, as well as longer training times.
They also used stochastic depth while we did not, which can in some cases contribute percent differences
in model accuracy~\citep{resmlp}. It is therefore possible that further hyperparameter tuning and more epochs for ConvMixers
could close the gap between the two architectures for large patches, \eg $p = 16$.

\textbf{ResMLPs.} Similarly to DeiT for ViT, we believe that ResMLP is the most relevant MLP-Mixer
variant to compare against. Unlike DeiT, we can compare against instances of ResMLP with similar patch size:
ResMLP-B24/8 has $p=8$ patches, and underperforms ConvMixer-1536/20 by 0.37\%,
despite having over twice the number of parameters; it also has similarly low throughput.
ConvMixer-768/32 also outperforms ResMLP-S12/8 for millions fewer parameters, but $3\times$ less throughput.

ResMLP did not significantly improve in terms of accuracy for halving the patch size from 16 to 8,
which shows that smaller patches do not always lead to better accuracy for a fixed architecture
and regularization strategy (\eg training a $p=8$ DeiT may be challenging).

\textbf{Swin Transformers.} While we intend to focus on the most basic isotropic, patch-based
architectures for fair comparisons with ConvMixer, it is also interesting to compare to a more complicated
model that is closer to state-of-the-art. For a similar parameter budget, ConvMixer is around 1.2-1.6\%
less accurate than the Swin Transformer, while also being 4-6$\times$ slower. However, considering
we did not attempt to tune or optimize our model in any way, we find it surprising that an exceedingly
simple patch-based model that uses only plain convolution does not lag too far behind Swin Transformer.

\textbf{Isotropic MobileNets.} These models are closest in design to ours, despite using a repeating
block that is substantially more complex than the ConvMixer one.
Despite this, for a similar number of parameters, we can get similar performance.
Notably, isotropic MobileNets seem to suffer less from larger patch sizes than ConvMixers,
which makes us optimistic that sufficient parameter tuning could lead to more performant large-patch ConvMixers. 

\looseness=-1
\textbf{Other models.} We included ViT and MLP-Mixer instances in our table, though they
are not competitive with ConvMixer, DeiT, or ResMLP, even though MLP-Mixer has comparable regularization
to ConvMixer. That is, ConvMixer seems to outperform MLP-Mixer and ViT, while being
closer to complexity to them in terms of design and training regime than the other competitors, DeiT and ResMLP.

\textbf{Kernel size.}
While we found some evidence that larger kernels are better on CIFAR-10, we wanted to see
if this finding transferred to ImageNet.
Consequently, we trained our best-performing model, ConvMixer-1536/20,
with kernel size $k = 3$ rather than $k = 9$.
This resulted in a decrease of 0.94\% top-1 accuracy,
which we believe is quite significant relative to the mere 2.2M additional parameters.
However, $k = 3$ is substantially faster than $k = 9$ for spatial-domain convolution;
we speculate that low-level optimizations could close the performance gap to some extent, \eg
by using implicit instead of explicit padding.
Since large-kernel convolutions throughout a model are unconventional, there has likely been low demand
for such optimizations.

\newpage
\section{Experiments on CIFAR-10}
\label{apx-cifar}

\textbf{Residual connections.}
We experimented with leaving out one, the other, or both residual connections
before settling on the current configuration,
and consequently chose to leave out the second residual connection.
Our baseline model without the connection achieves 95.88\% accuracy,
while including the connection reduces it to 94.78\%.
Surprisingly, we see only a $0.31\%$ decrease in accuracy for \emph{removing all residual connections}.
We acknowledge that these findings for residual connections may not generalize to deeper ConvMixers
trained on larger data sets.

\begin{table}[h]
\begin{center}
\begin{tabularx}{0.5\textwidth}{l|Y}
\toprule
\multicolumn{2}{c}{Ablation of ConvMixer-256/8 on CIFAR-10} \\
\midrule
\thead{Ablation} &
\thead{CIFAR-10\\Acc. (\%)}  \\
\midrule
Baseline & 95.88 \\
\midrule
-- Residual in Eq.~\ref{eq-depthwise} & 95.57 \\
+ Residual in Eq.~\ref{eq-pointwise} & 94.78 \\
BatchNorm $\to$ LayerNorm & 94.44 \\
GELU $\to$ ReLU & 95.51 \\
\midrule
-- Mixup and CutMix & 95.92 \\
-- Random Erasing & 95.24 \\
-- RandAug & 92.86 \\
-- Random Scaling & 86.24 \\
-- Gradient Norm Clipping & 86.33 \\
\bottomrule
\end{tabularx}
\caption{Small ablation study of training a ConvMixer-256/8 on CIFAR-10.}
\label{cifar-ablate}
\end{center}
\end{table}

\textbf{Normalization.}
Our model is conceptually similar to the vision transformer and MLP-Mixer, both
of which use LayerNorm instead of BatchNorm.
We attempted to use LayerNorm instead,
and saw a decrease in performance of around 1\% as well as slower convergence (see Table~\ref{cifar-ablate}).
However, this was for a relatively shallow model, and we cannot guarantee that LayerNorm
would not hinder ImageNet-scale models to an even larger degree.
We note that the authors of ResMLP also saw a relatively small increase in accuracy
for replacing LayerNorm with BatchNorm, but for a larger-scale experiment~\citep{resmlp}.
We conclude that BatchNorm is no more crucial to our architecture
than other regularizations or parameter settings (\eg kernel size).

Having settled on an architecture, we proceeded to adjust its parameters $h, d, p, k$
as well as weight decay on CIFAR-10 experiments.
(Initially, we took the unconventional approach of excluding weight decay
since we were already using strong regularization in the form of RandAug and mixup.)
We acknowledge that tuning our architecture on CIFAR-10 does not necessarily generalize
to performance on larger data sets, and that this is a limitation of our study.

\subsection{Results}
ConvMixers are quite performant on CIFAR-10,
easily achieving $>91\%$ accuracy for as little as $100,000$
parameters, or $>96\%$ accuracy for only $887,000$ parameters (see Table~\ref{tab-cifar}).
With additional refinements \eg a more expressive classifier or bottlenecks,
we think that ConvMixer could be even more competitive.
For all experiments, we trained for 200 epochs on CIFAR-10 with
RandAug, mixup, cutmix, random erasing, gradient norm clipping,
and the standard augmentations in \texttt{timm}.
We remove some of these augmentations in Table~\ref{cifar-ablate},
finding that RandAug and random scaling (``default'' in \texttt{timm})
are very important, each accounting for over $3\%$ of the accuracy.

\textbf{Scaling ConvMixer.} We adjusted the hidden dimension $h$ and the depth $d$,
finding that deeper networks take longer to converge while wider networks
converge faster. That said, increasing the width or the depth is an effective
way to increase accuracy; a doubling of depth incurs less compute
than a doubling of width. The number of parameters in a ConvMixer
is given exactly by:%
\begin{equation}
\mathsf{\#params} = h[d(k^2 + h + 6) + \cin p^2 + n_{\mathsf{classes}} + 3] + n_{\mathsf{classes}},
\end{equation}%
including affine scaling parameters in BatchNorm layers, convolutional kernels,
and the classifier.

\textbf{Kernel size.}
We initially hypothesized that large kernels would be important for ConvMixers,
as they would allow the mixing of distant spatial information similarly
to unconstrained MLPs or self-attention layers.
We tried to investigate the effect of kernel size on CIFAR-10:
we fixed the model to be a ConvMixer-256/8,
and increased the kernel size by 2s from 3 to 15.

Using a kernel size of 3, the ConvMixer only achieves 93.61\% accuracy.
Simply increasing it to 5 gives an additional 1.50\% accuracy,
and further to 7 an additional 0.61\%.
The gains afterwards are relatively marginal, with kernel size 15 giving
an additional 0.28\% accuracy.
It could be that with more training iterations or more regularization,
the effect of larger kernels would be more pronounced.
Nonetheless, we concluded that ConvMixers benefit from larger-than-usual kernels,
and thus used kernel sizes 7 or 9 in most of our later experiments.

It is conventional wisdom that large-kernel convolutions can be ``decomposed'' into
stacked small-kernel convolutions with activations between them, and it is
therefore standard practice to use $k= 3$ convolutions, stacking more of them
to increase the receptive field size with additional benefits from nonlinearities.
This raises a question: is the benefit of larger kernels in ConvMixer actually
better than simply increasing the depth with small kernels?
First, we note that deeper networks are generally harder to train,
so by increasing the kernel size independently of the depth,
we may recover some of the benefits of depth without making it harder for
signals to ``propagate back'' through the network.
To test this, we trained a ConvMixer-256/10 with $k = 3$ (698K parameters)
in the same setting as a ConvMixer-256/8 with $k = 9$ (707K parameters),
\ie we increased depth in a small-kernel model to roughly match the parameters
of a large-kernel model.
The ConvMixer-256/10 achieved 94.29\% accuracy (1.5\% less),
which provides more evidence for the importance of larger kernels
in ConvMixers.
Next, instead of fixing the parameter budget, we tripled the depth (using the intuition
that 3 stacked $k=3$ convolutions have the receptive field of a $k=9$ convolution),
giving a ConvMixer-256/24 with 1670K parameters,
and got 95.16\% accuracy, \ie still less.

\looseness=-1
\textbf{Patch size.}
CIFAR-10 inputs are so small that we initially only used $p = 1$, \ie
the patch embedding layer does little more than compute $h$ linear combinations of the input image.
Using $p=2$, we see a reduction in accuracy of about 0.80\%;
this is a worthy tradeoff in terms of training and inference time.
Further increasing the patch size leads to rapid decreases in accuracy,
with only 92.61\% for $p = 4$.

\looseness=-1
Since the ``internal resolution'' is decreased by a factor of $p$ when increasing the patch size,
we assumed that larger kernels would be less important for larger $p$.
We investigated this by again increasing the kernel size from 3 to 11 for  ConvMixer-256/8 with $p=2$:
however, this time, the improvement going from 3 to 5 is only 1.13\%,
and larger kernels than 5 provide only marginal benefit.

\textbf{Weight decay.} We did many of our initial experiments with minimal weight decay.
However, this was not optimal: by tuning weight decay, we can get an additional 0.15\%
of accuracy for no cost. Consequently, we used weight decay (without tuning) for
our larger-scale experiments on ImageNet.

\begin{table}[hb!]
\begin{center}
\begin{tabularx}{\textwidth}{Y|Y|Y|Y|Y|Y|Y}
\toprule
\multicolumn{7}{c}{Tiny \textbf{ConvMixers} trained on CIFAR-10.} \\
\midrule
\thead{Width\\$h$} &
\thead{Depth\\ $d$} &
\thead{Patch\\Size $p$} &
\thead{Kernel \\ Size $k$} &
\thead{\# Params\\($\times10^3$)} &
\thead{Weight\\Decay} &
\thead{CIFAR-10\\Acc. (\%)}  \\
\midrule
128 &  4 & 1 & 8 & 103  & 0 & 91.26 \\
128 &  8 & 1 & 8 & 205  & 0 & 93.83 \\
128 & 12 & 1 & 8 & 306  & 0 & 94.83 \\
256 &  4 & 1 & 8 & 338  & 0 & 93.37 \\
256 &  8 & 1 & 8 & 672  & 0 & 95.60 \\
256 & 12 & 1 & 8 & 1006 & 0 & 96.39 \\
256 & 16 & 1 & 8 & 1339 & 0 & 96.74 \\
256 & 20 & 1 & 8 & 1673 & 0 & 96.67 \\
\midrule
\multicolumn{7}{l}{$\downarrow$ Kernel adjustments} \\
\midrule
256 &  8 & 1 & 3  & 559   & 0 & 93.61 \\
256 &  8 & 1 & 5  & 592   & 0 & 95.19 \\
256 &  8 & 1 & 7  & 641   & 0 & 95.80 \\
256 &  8 & 1 & 9  & 707   & 0 & 95.88 \\
256 &  8 & 1 & 11 & 788   & 0 & 95.70 \\
256 &  8 & 1 & 13 & 887   & 0 & 96.04 \\
256 &  8 & 1 & 15 & 1001  & 0 & 96.08 \\
\midrule
\multicolumn{7}{l}{$\downarrow$ Patch adjustments} \\
\midrule
256 &  8 & 2 & 9 & 709  & 0 & 95.00 \\
256 &  8 & 4 & 9 & 718  & 0 & 92.61 \\
256 &  8 & 8 & 9 & 755  & 0 & 85.57 \\
\midrule
\multicolumn{7}{l}{$\downarrow$ Weight decay adjustments} \\
\midrule
256 &  8 & 1 & 9  & 707  & $1\times10^{-1}$ & 95.88 \\
256 &  8 & 1 & 9  & 707  & $1\times10^{-2}$ & 96.03 \\
256 &  8 & 1 & 9  & 707  & $1\times10^{-3}$ & 95.76 \\
256 &  8 & 1 & 9  & 707  & $1\times10^{-4}$ & 95.63 \\
256 &  8 & 1 & 9  & 707  & $1\times10^{-5}$ & 95.88 \\
\midrule
\multicolumn{7}{l}{$\downarrow$ Kernel size adjustments when $p=2$} \\
\midrule
256 &  8 & 2 & 3 & 561  & 0 & 94.08 \\
256 &  8 & 2 & 5 & 594  & 0 & 95.21 \\
256 &  8 & 2 & 7 & 643  & 0 & 95.35 \\
256 &  8 & 2 & 9 & 709  & 0 & 95.00 \\
256 &  8 & 2 & 11 & 791  & 0 & 95.14 \\
\midrule
\multicolumn{7}{l}{$\downarrow$ Adding weight decay to the above} \\
\midrule
256 &  8 & 2 & 3 & 561  & $1\times10^{-2}$ & 94.69 \\
256 &  8 & 2 & 5 & 594  & $1\times10^{-2}$ & 95.26 \\
256 &  8 & 2 & 7 & 643  & $1\times10^{-2}$ & 95.25 \\
256 &  8 & 2 & 9 & 709  & $1\times10^{-2}$ & 95.06 \\
256 &  8 & 2 & 11 & 791 & $1\times10^{-2}$ & 95.17 \\
\bottomrule
\end{tabularx}
\caption{An investigation of ConvMixer design parameters $h, d, p, k$ and weight decay
on CIFAR-10}
\vspace*{-1em}
\label{tab-cifar}
\end{center}
\end{table}

\clearpage

\section{Weight Visualizations}
\begin{figure}[h!]
	\centering
	\includegraphics[width=0.8\textwidth]{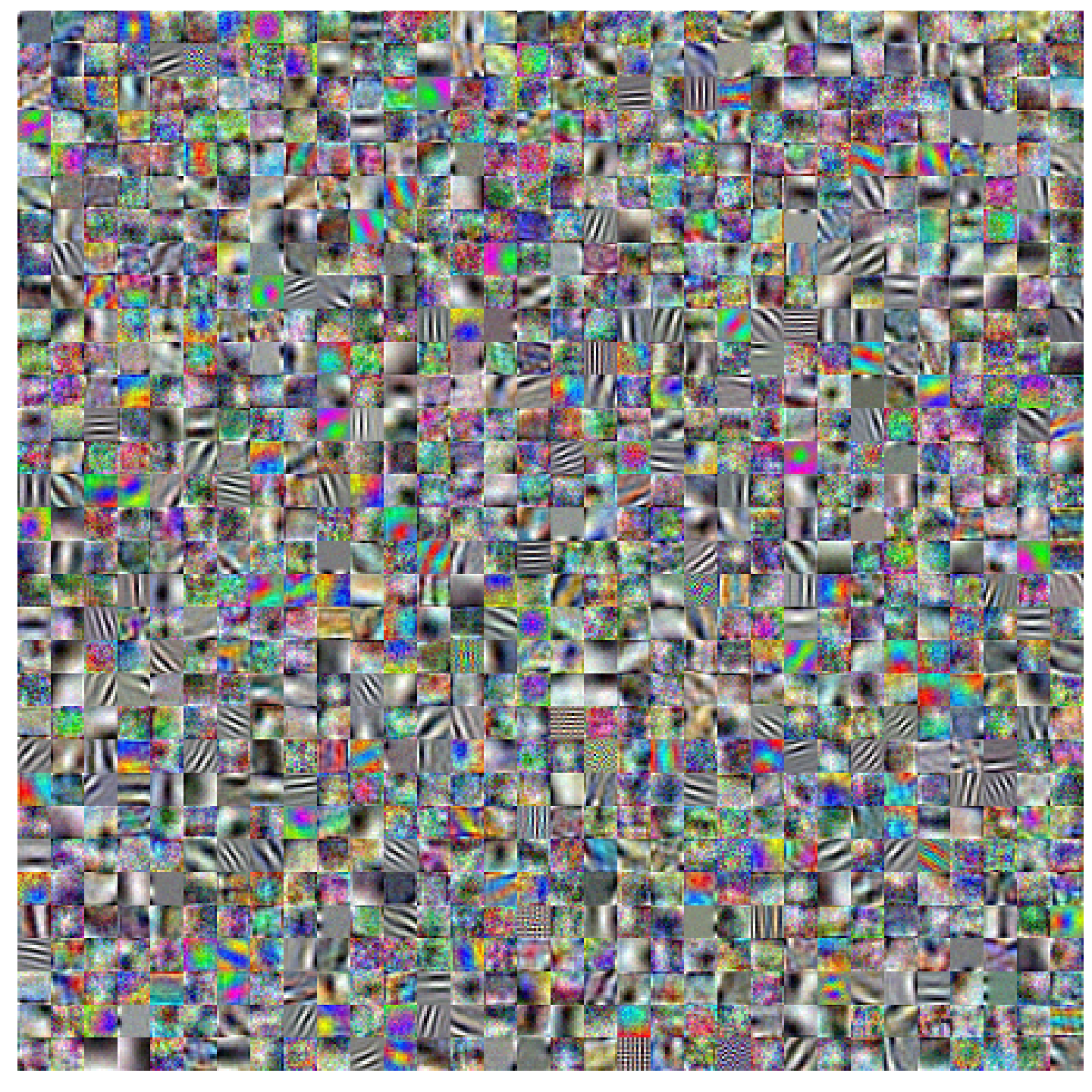}
	\caption{Patch embedding weights for a ConvMixer-1024/20 with patch size 14 (see Table~\ref{models}).}
	\label{fig-cmixp14}
\end{figure}

\begin{figure}[h]
	\centering
	\includegraphics[width=0.4\textwidth]{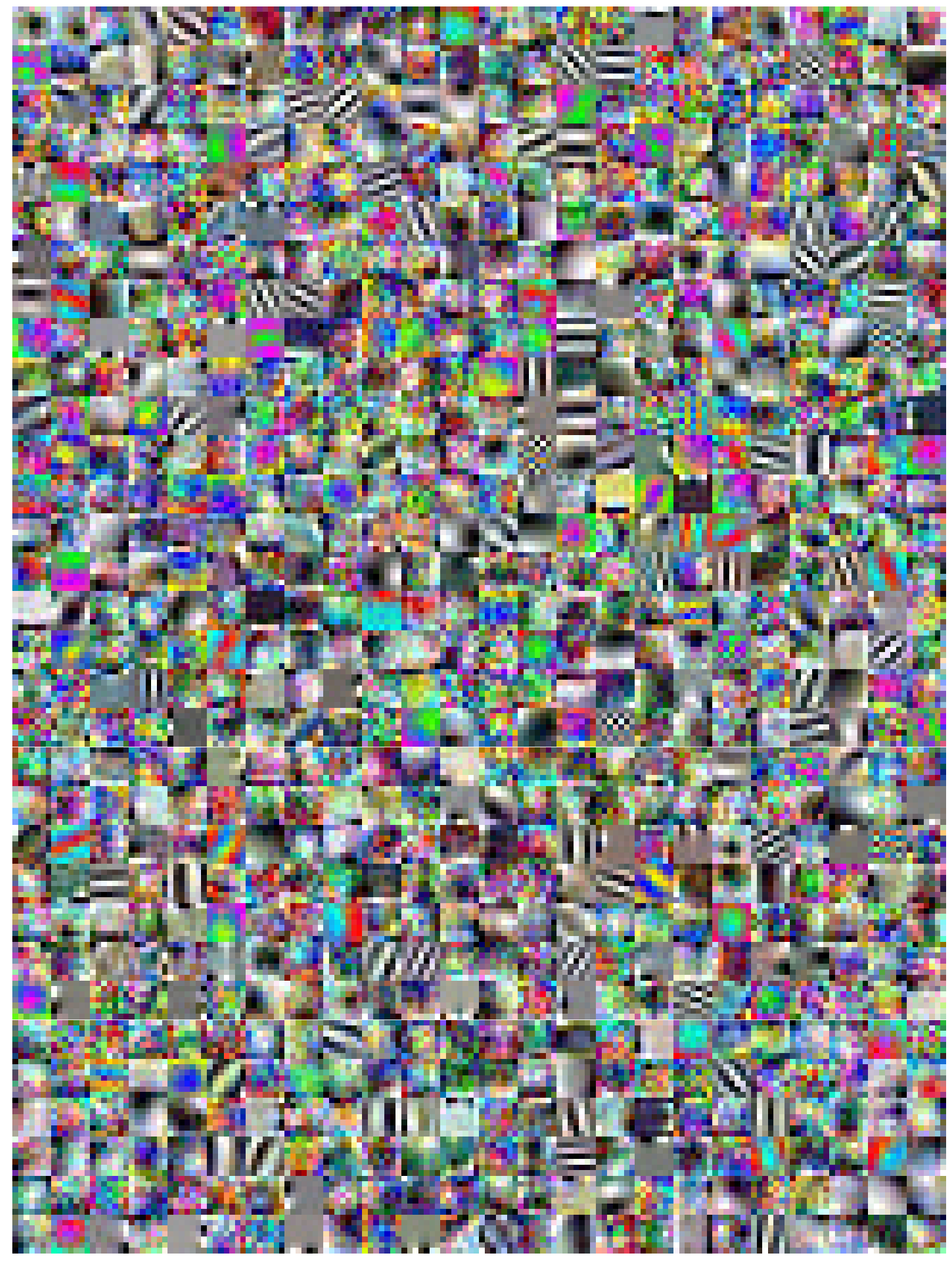}
	\caption{Patch embedding weights for a ConvMixer-768/32 with patch size 7 (see Table~\ref{models}).}
	\label{fig-cmixp7}
\end{figure}

\newpage
\begin{figure}[h!]
	\centering
	\subfigure[Layer 1]{\includegraphics[width=0.3\textwidth]{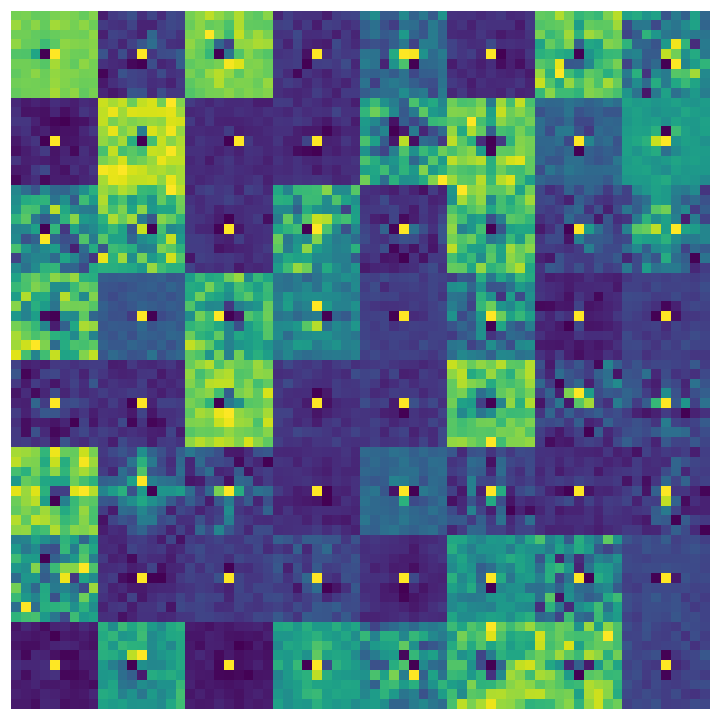}}
	\subfigure[Layer 4]{\includegraphics[width=0.3\textwidth]{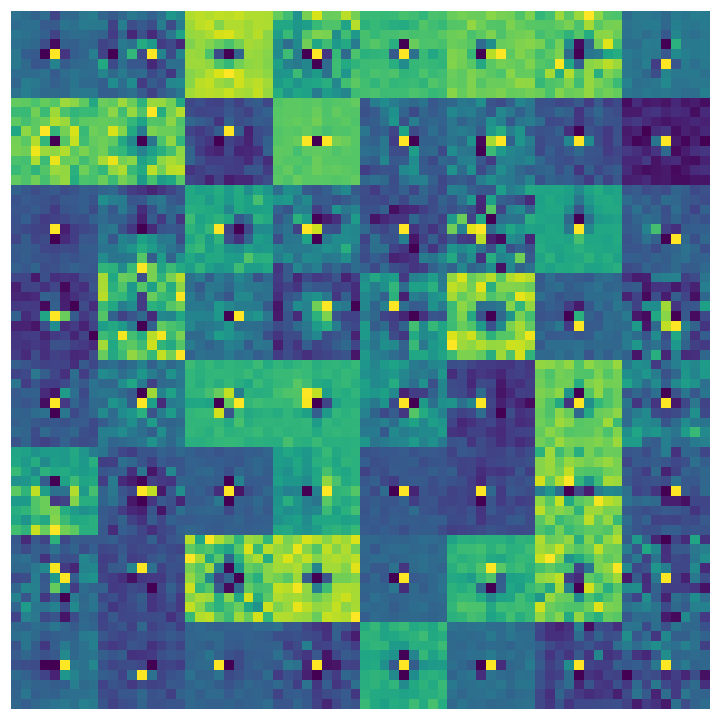}}
	\subfigure[Layer 6]{\includegraphics[width=0.3\textwidth]{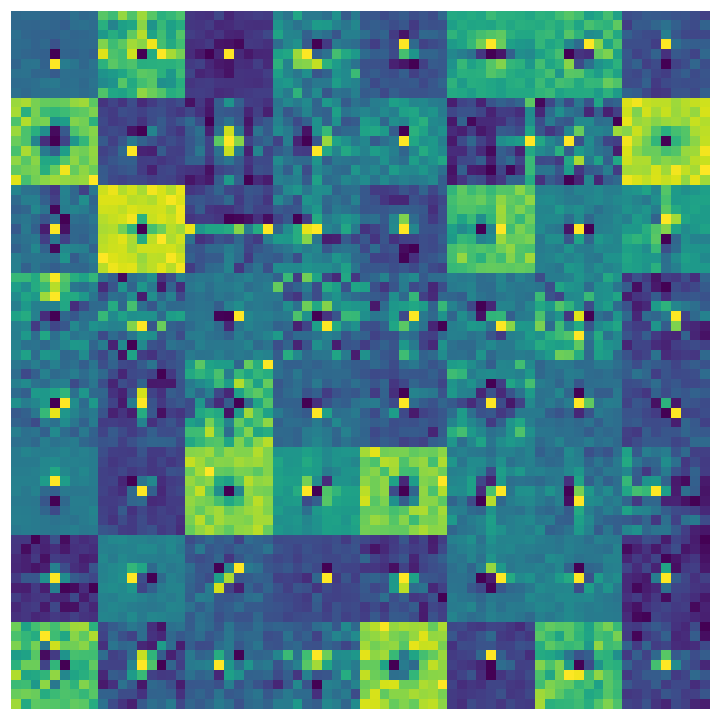}}
	\subfigure[Layer 9]{\includegraphics[width=0.3\textwidth]{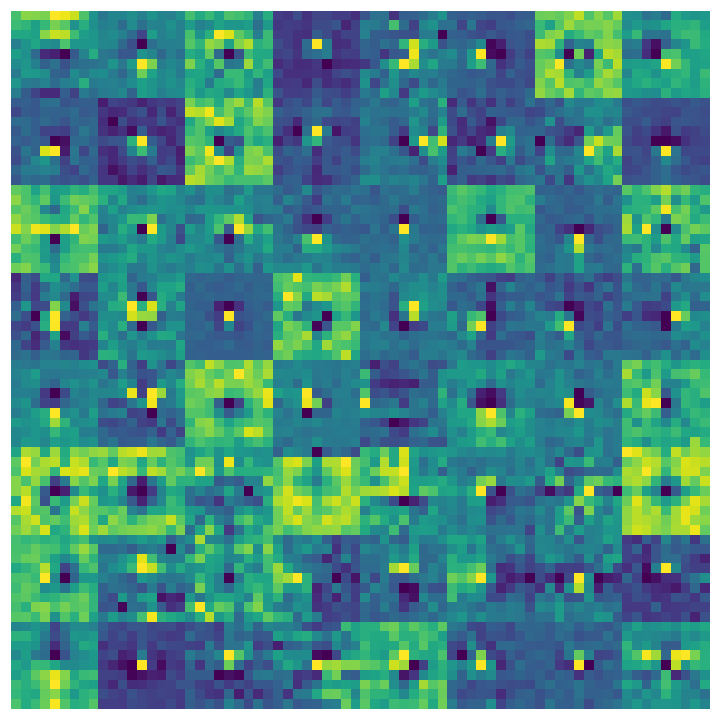}}
	\subfigure[Layer 11]{\includegraphics[width=0.3\textwidth]{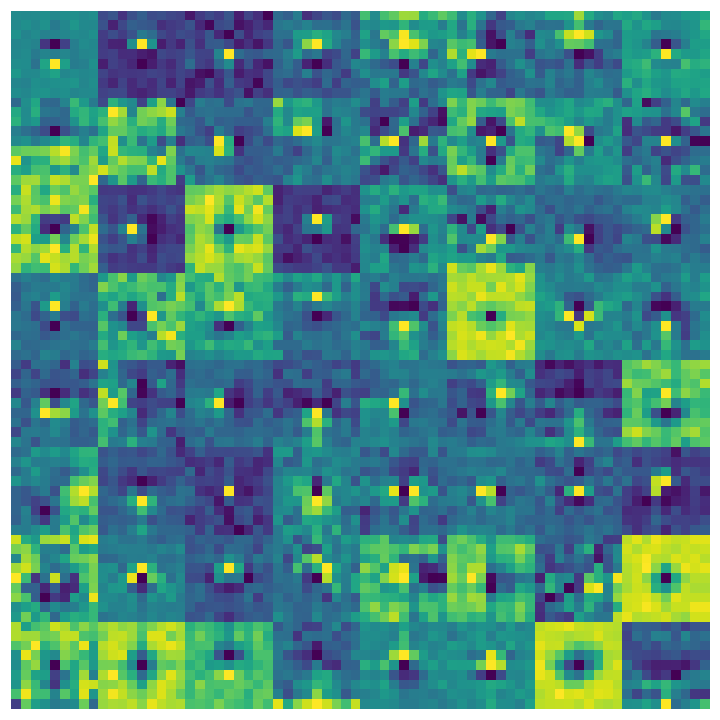}}
	\subfigure[Layer 13]{\includegraphics[width=0.3\textwidth]{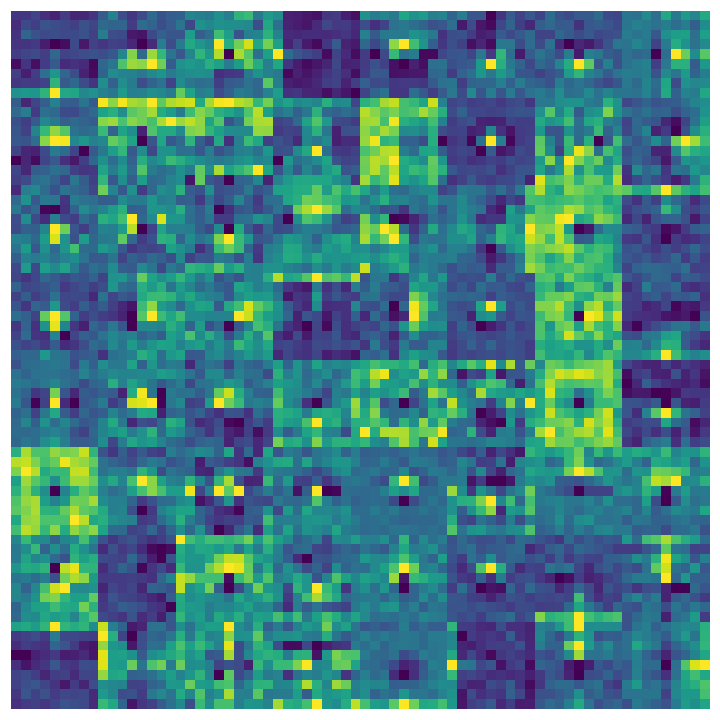}}
	\subfigure[Layer 15]{\includegraphics[width=0.3\textwidth]{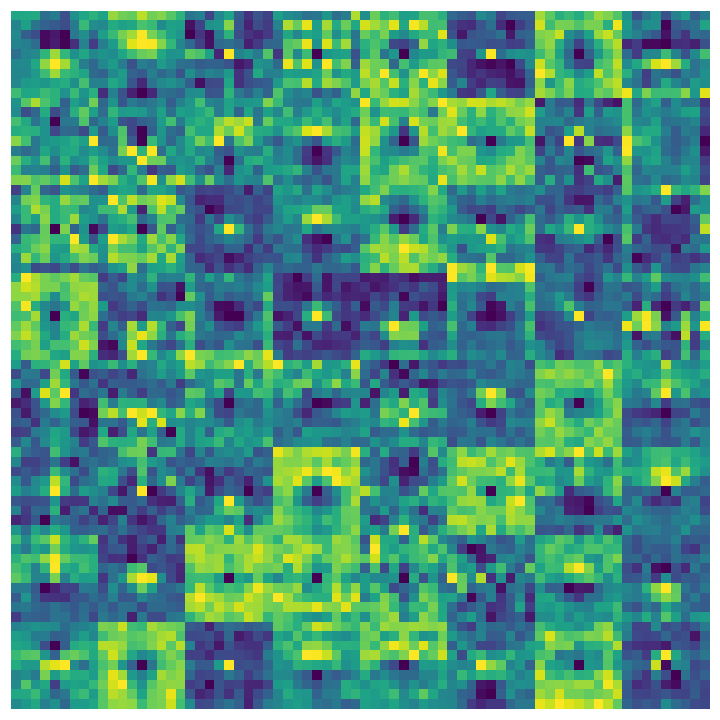}}
	\subfigure[Layer 17]{\includegraphics[width=0.3\textwidth]{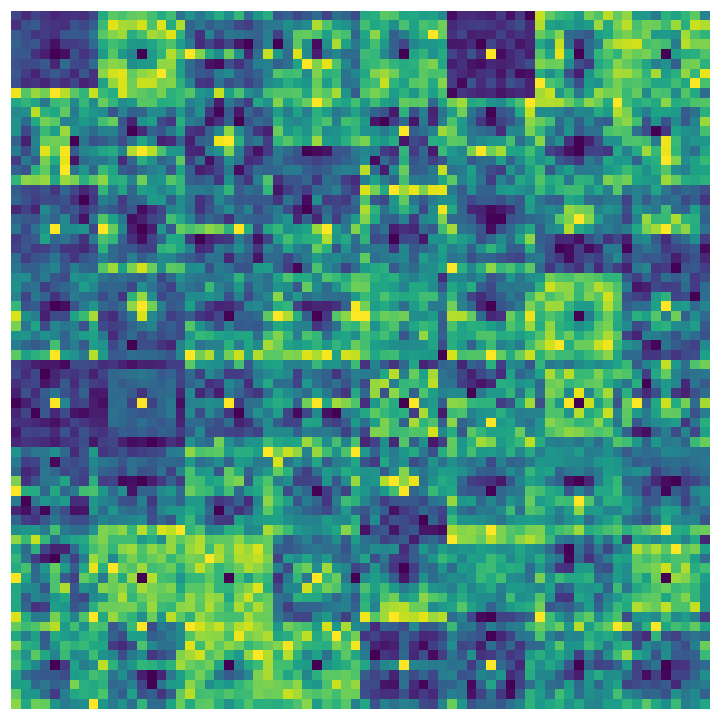}}
	\subfigure[Layer 19]{\includegraphics[width=0.3\textwidth]{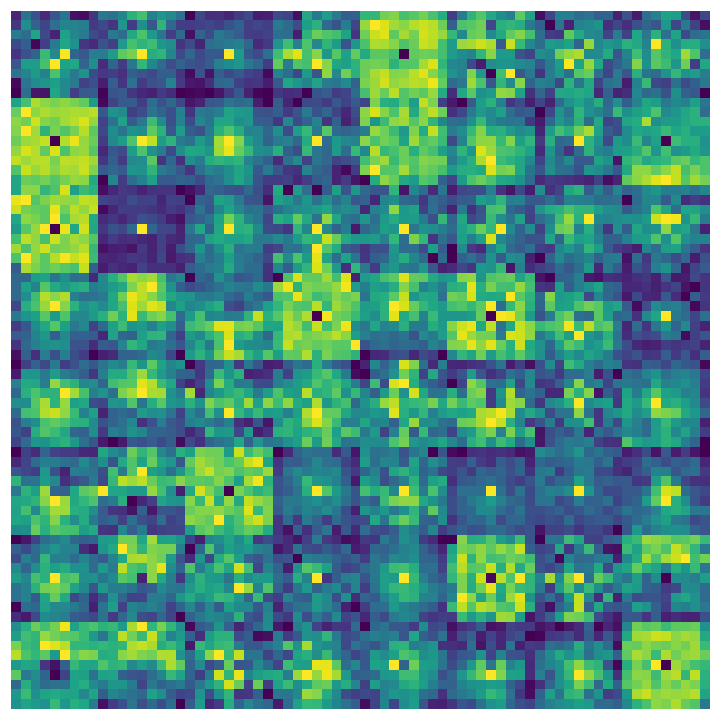}}
	\caption{Random subsets of 64 depthwise convolutional kernels from progressively deeper layers of ConvMixer-1536/20 (see Table~\ref{convmixers}).}
	\label{fig-cmixh}
\end{figure}

In Figure~\ref{fig-cmixp14} and \ref{fig-cmixp7}, we visualize the (complete) weights
of the patch embedding layers of a ConvMixer-1536/20 with $p=14$ and a ConvMixer-768/32 with $p=7$, respectively.
Much like \cite{isomobile}, the layer consists of Gabor-like filters as well as ``colorful globs'' or
rough edge detectors.
The filters seem to be more structured than those learned by MLP-Mixer~\citep{mixer};
also unlike MLP-Mixer, the weights look much the same going from $p=14$ to $p=7$: the latter
simply looks like a downsampled version of the former.
It is unclear, then, why we see such a drop in accuracy for larger patches.
However, some of the filters essentially look like noise, maybe suggesting a need for more regularization
or longer training, or even more data. Ultimately, we cannot read too much into the learned representations here.

In Figure~\ref{fig-cmixh}, we plot the hidden convolutional kernels for successive layers of a ConvMixer.
Initially, the kernels seem to be relatively small, but make use of their allowed full size in later layers;
there is a clear hierarchy of features as one would expect from a standard convolutional architecture.
Interestingly, \cite{resmlp} saw a similar effect for ResMLP, where earlier layers look like small-kernel convolution,
while later layers were more diffuse, despite these layers being representated by an unconstrained matrix multiplication
rather than convolution.

\newpage

\section{Implementation}
\label{apx-implementation}

\begin{figure}[h]
{\footnotesize
\begin{minted}[linenos]{python}
import torch.nn as nn

class Residual(nn.Module):
    def __init__(self, fn):
        super().__init__()
        self.fn = fn

    def forward(self, x):
        return self.fn(x) + x

def ConvMixer(dim, depth, kernel_size=9, patch_size=7, n_classes=1000):
    return nn.Sequential(
        nn.Conv2d(3, dim, kernel_size=patch_size, stride=patch_size),
        nn.GELU(),
        nn.BatchNorm2d(dim),
        *[nn.Sequential(
                Residual(nn.Sequential(
                    nn.Conv2d(dim, dim, kernel_size, groups=dim, padding="same"),
                    nn.GELU(),
                    nn.BatchNorm2d(dim)
                )),
                nn.Conv2d(dim, dim, kernel_size=1),
                nn.GELU(),
                nn.BatchNorm2d(dim)
        ) for i in range(depth)],
        nn.AdaptiveAvgPool2d((1,1)),
        nn.Flatten(),
        nn.Linear(dim, n_classes)
    )
\end{minted}
}
	\vspace*{-1em}
	\caption{A more readable PyTorch~\citep{pytorch} implementation of ConvMixer, where $h$ = \texttt{dim}, $d$ = \texttt{depth},
	$p$ = \texttt{patch\_size}, $k$ = \texttt{kernel\_size}.}
	\label{readable}
\end{figure}

\begin{figure}[h]
\centering
{\footnotesize
\begin{minted}[linenos,breaklines]{python}
def ConvMixer(h,d,k,p,n):
 S,C,A=Sequential,Conv2d,lambda x:S(x,GELU(),BatchNorm2d(h))
 R=type('',(S,),{'forward':lambda s,x:s[0](x)+x})
 return S(A(C(3,h,p,p)),*[S(R(A(C(h,h,k,groups=h,padding=k//2))),A(C(h,h,1))) for i in range(d)],AdaptiveAvgPool2d(1),Flatten(),Linear(h,n))
\end{minted}
}
	\vspace*{-1em}
	\caption{An implementation of our model in less than 280 characters, in case you happen to know of any means of disseminating information that could benefit from such a length.\newline
	All you need to do to run this is \texttt{from torch.nn import *}.}
	\label{golfing}
	\vspace{1em}
\end{figure}

This section presents an expanded (but still quite compact)
version of the terse ConvMixer implementation that we presented in the paper.
The code is given in Figure~\ref{readable}.
We also present an \emph{even more terse}
implementation in Figure~\ref{golfing}, which to the best
of our knowledge is the first model that achieves the elusive
dual goals of $80\%+$ ImageNet top-1 accuracy while also fitting into a tweet.

\end{document}